%% file: 0main.tex
\documentclass{article} 
\usepackage{iclr2027_conference,times}

\input{math_commands.tex}

\usepackage{hyperref}
\usepackage{url}
\usepackage{multirow}
\usepackage{multicol}
\usepackage{xspace}
\usepackage{wrapfig}
\usepackage{booktabs}
\usepackage{graphicx}
\usepackage{tcolorbox}
\tcbuselibrary{skins}
\usepackage[framemethod=TikZ]{mdframed}
\usepackage{algorithm}
\usepackage{algpseudocode}
\usepackage[accsupp]{axessibility}
\usepackage{enumitem}
\usepackage{fvextra}
\newcommand{\ours}{\texttt{Clue-OPSD}\xspace}

\title{Where to Look Matters: On-Policy Self-Distillation for Long-Video Understanding}

\author{
Kaishen Wang$^{1}$,
Dongdi Zhao,
Yijun Liang$^{1}$,
Dingqiang Ye$^{2}$,
Ruibo Chen$^{1}$,
{Heng Huang}$^{1}$,
{Di Fu}
\\[6pt]
\multicolumn{1}{c}{
$^{1}$University of Maryland, College Park,
$^{2}$Johns Hopkins University
}
\\[2pt]
\multicolumn{1}{c}{
\texttt{kaishen@umd.edu}
}
}

\iclrfinalcopy 
\begin{document}

\maketitle


\lhead{Preprint}

\begin{abstract}
Vision-language models (VLMs) have made substantial progress in long-video understanding, with standard backbone models typically answering questions from frames sampled across the full video. However, as videos become longer, the full-video context inevitably contains more question-irrelevant temporal content, which can distract the model from the evidence needed to answer a specific question. We empirically find that focusing the visual input on short annotated clue intervals containing question-relevant evidence consistently improves prediction accuracy across model scales compared with using the corresponding full videos, while requiring fewer input frames. Based on this finding, we introduce \textbf{\ours}, a clue-privileged on-policy self-distillation framework for long-video understanding. During training, a full-video student learns from a self-teacher conditioned on the corresponding clue interval by aligning their next-token distributions along student-generated trajectories. \ours thus uses clue intervals as privileged supervision without relying on ground-truth answer labels, while requiring no clue annotations or additional modules at inference time. Extensive experiments across multiple long-video understanding benchmarks and Qwen3.5 model scales demonstrate consistent improvements over the corresponding backbone models and strong performance against supervised post-training baselines.
\end{abstract}

\input{1intro}
\input{2related}
\input{3method}
\input{4exp}
\input{5con}

\bibliography{iclr2027_conference}
\bibliographystyle{iclr2027_conference}

\clearpage
\input{6appendix}

\end{document}

%% file: math_commands.tex
\usepackage{amsmath,amsfonts,bm}

\def\eqref#1{equation~\ref{#1}}

\def\1{\bm{1}}

\DeclareMathAlphabet{\mathsfit}{\encodingdefault}{\sfdefault}{m}{sl}
\SetMathAlphabet{\mathsfit}{bold}{\encodingdefault}{\sfdefault}{bx}{n}



%% file: 1intro.tex
\section{Introduction}

Recent advances in foundation vision-language models (VLMs) have substantially improved video understanding and enabled models to process increasingly long visual contexts~\citep{chen2024expanding,bai2025qwen3,zhu2025internvl3,qwen35blog}. For long-video question answering, standard backbone models typically sample frames across the full video and jointly process them with the textual query~\citep{ranasinghe2025understanding,shu2025video,li2026videochat,lin2026unleashing}. With stronger backbones and larger visual context windows, this simple full-video paradigm has become increasingly effective.

\begin{figure}
    \centering
    \includegraphics[width=0.85\linewidth]{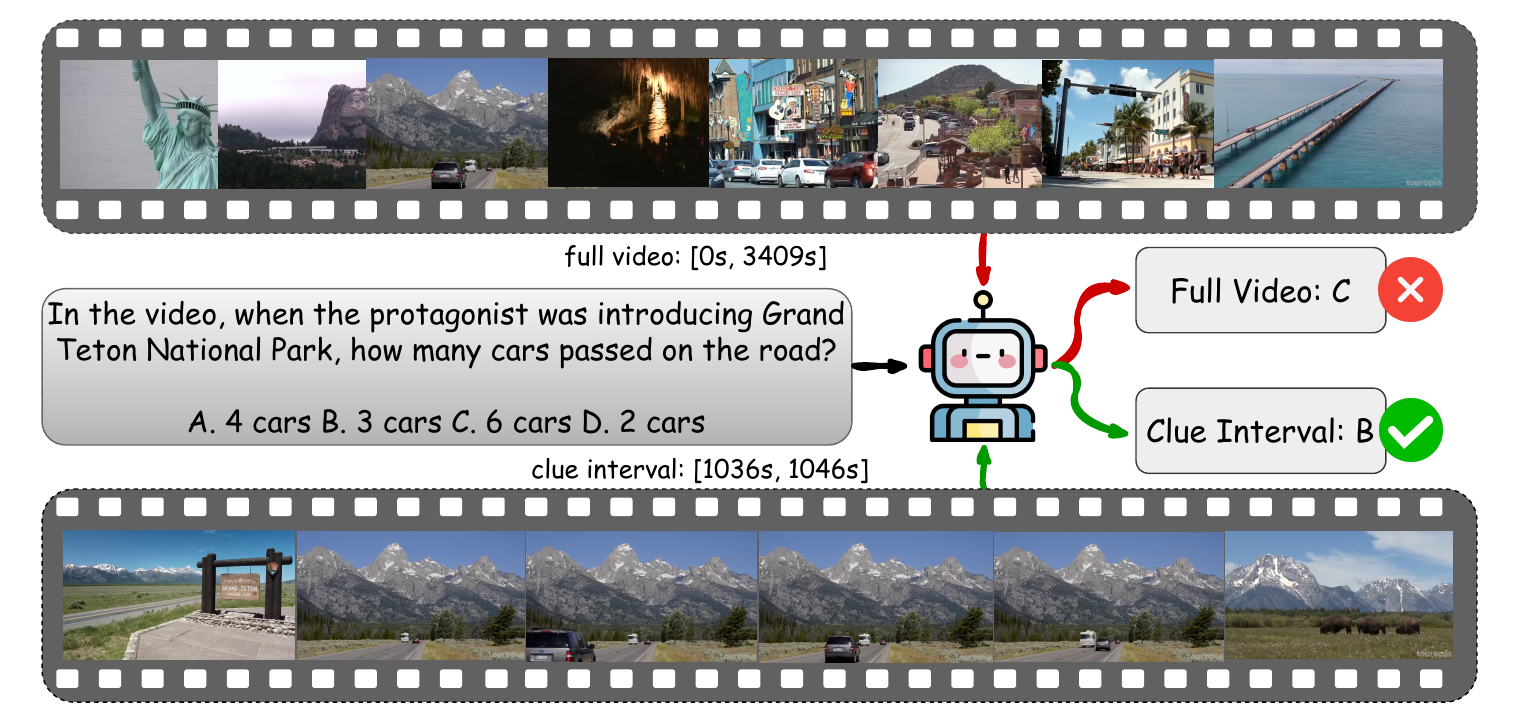}
    \vspace{-1mm}
    \caption{Illustration of the effect of clue intervals in long-video question answering. The model fails when processing frames sampled from the full 3,409-second video but answers correctly when conditioned on the corresponding 10-second clue interval containing the question-relevant evidence.}
    \label{fig:example_local_full}
    \vspace{-1mm}
\end{figure}

However, as videos become longer, the full-video context inevitably contains more temporal content that is unrelated to a particular question. The model must therefore identify the useful evidence from a large amount of surrounding visual information before making a prediction, and relevant events can easily be obscured by unrelated content distributed across the video. Figure~\ref{fig:example_local_full} illustrates this issue with a representative example. When the model processes frames sampled from the full 3,409-second video, it predicts an incorrect answer. In contrast, when the visual input is restricted to the corresponding 10-second clue interval containing the relevant event, the same model answers the question correctly.

We further examine this phenomenon systematically using CG-Bench~\citep{chen2025cg}, which provides annotated temporal clue intervals for individual questions. Under the same frames-per-second (FPS) and maximum-frame constraints, we compare the same VLM using frames sampled from the full video with frames sampled only from the corresponding clue interval. As shown in Table~\ref{tab:clue_observation}, across different Qwen3.5 model scales, focusing the visual input on short annotated clue intervals containing question-relevant evidence consistently improves prediction accuracy compared with using the corresponding full videos, while requiring fewer input frames. These results show that, for long-video question answering, the way relevant temporal evidence is presented to the model can substantially affect prediction quality.

Based on this finding, we introduce \textbf{\ours}, a clue-privileged on-policy self-distillation (OPSD) framework for long-video understanding. During training, \ours constructs two asymmetric visual conditions from the same video: a student operates on the full-video input, while an exponential moving average (EMA) self-teacher observes only the annotated clue interval associated with the current question. The student first generates an on-policy trajectory, and both branches are evaluated along the same student-generated prefixes. We then align their next-token predictive distributions, allowing the full-video student to learn from a teacher conditioned on the corresponding clue interval. Importantly, the clue annotations are used only to construct the privileged teacher input during training. \ours does not use ground-truth answer labels as supervision and does not require a separately pretrained or larger teacher model. At inference time, both the clue annotations and EMA teacher are removed, and the trained student directly processes the full video using the standard VLM inference pipeline, without additional temporal localization or auxiliary inference-time modules.

We evaluate \ours across multiple long-video understanding benchmarks using Qwen3.5 models at different scales. Across model sizes and evaluation settings, \ours consistently improves over the corresponding backbone models and achieves strong performance against supervised post-training baselines. The contributions of this paper are summarized as follows:
\begin{itemize}
    \item We show that focusing the visual input on short annotated clue intervals containing question-relevant evidence can consistently outperform full-video input across different VLM scales, even with fewer input frames.
    \item We propose \ours, a clue-privileged on-policy self-distillation framework that uses clue intervals as privileged visual supervision for a full-video student without relying on ground-truth answer labels or an external teacher model.
    \item We demonstrate consistent improvements across multiple Qwen3.5 model scales and long-video understanding benchmarks while retaining the standard full-video VLM inference pipeline without additional inference-time modules.
\end{itemize}

%% file: 2related.tex
\section{Related Work}

\subsection{Long-Video Understanding}

With the rapid progress of foundation vision-language models (VLMs)~\citep{chen2024expanding,bai2025qwen3,zhu2025internvl3,qwen35blog,wang2026mitigating,wang2026unsafe}, long-video understanding has improved substantially. Standard backbone models typically sample frames across the full video and process them together with the textual query~\citep{ranasinghe2025understanding,shu2025video,li2026videochat,lin2026unleashing}. This paradigm enables direct reasoning over increasingly long temporal contexts, while recent work has further explored more efficient representations and compression strategies for long-video inputs~\citep{shu2025video,li2026videochat,zhang2025flash}.

More recently, \textit{thinking-with-videos} methods have explored a more active inference paradigm by searching for, retrieving, or temporally grounding question-relevant video segments and iteratively inspecting them during reasoning~\citep{wang2024videoagent,yang2025vca,yuan2025videodeepresearch,liu2026videotemp,yang2026longvt,zhang2026deep}. While effective, these approaches typically require multiple rounds of temporal search, segment inspection, or tool interaction, leading to more complex and time-consuming inference. In contrast, our work uses annotated temporal clue intervals only during training as privileged visual supervision, while retaining the standard full-video VLM pipeline at inference time.

\subsection{On-Policy Distillation and Self-Distillation}

On-policy distillation (OPD) trains a student on trajectories sampled from its current policy while using a teacher to provide token-level distribution supervision~\citep{agarwal2024policy}. By matching teacher and student predictions on student-generated trajectories, OPD reduces the mismatch between training-time supervision and the states encountered during autoregressive inference. Recent studies have extended this paradigm through different teacher constructions, conditioning contexts, and supervision strategies~\citep{yu2026weak,tan2026self,sang2026policy,song2026survey,ye2026policy,hou2026dash}. In particular, on-policy self-distillation (OPSD)~\citep{zhao2026self} allows the same underlying model to serve as both teacher and student under asymmetric conditioning, where privileged information available only to the teacher provides the supervision signal.

This idea has recently been extended to multimodal learning. Vision-OPD~\citep{yuan2026vision} uses evidence-centered image crops as privileged visual context for a full-image student, while Visual-OPSD~\citep{li2026visual} and Visual Contrastive Self-Distillation~\citep{liang2026visual} explore cross-modal and contrastive forms of asymmetric visual supervision. Video-OPD~\citep{li2026video} applies on-policy distillation to temporal video grounding. Our work focuses instead on long-video understanding, where question-relevant temporal clue intervals are used as privileged visual context to supervise a student operating directly on the full video.

%% file: 3method.tex
\section{Method}

\subsection{Preliminaries}

\paragraph{Video Question Answering.}
Given a raw video $\mathcal{V}$ and a question $q$, we first convert the video into a sequence of temporally sampled frames:
\begin{equation}
V=\{v_1,\ldots,v_N\},
\end{equation}
where $N$ denotes the number of sampled frames used as visual input to the model. A vision-language model (VLM) parameterized by $\theta$ then autoregressively generates an answer $y=(y_1,\ldots,y_T)$ according to:
\begin{equation}
p_{\theta}(y\mid V,q)
=
\prod_{t=1}^{T}
p_{\theta}
\left(
y_t
\mid
V,q,y_{<t}
\right),
\end{equation}
where $y_{<t}$ denotes the previously generated tokens.

\paragraph{On-Policy Distillation.}
On-policy distillation (OPD) performs token-level knowledge distillation along trajectories generated by the current student policy. Given an input $x$, the student first samples an autoregressive response:
\begin{equation}
\hat{y}\sim p_{\theta_S}(\cdot\mid x),
\end{equation}
where $\theta_S$ denotes the parameters of the student model. The sampled response is then used as the shared autoregressive trajectory for both the student and teacher. At each decoding step $t$, conditioned on the same prefix $\hat{y}_{<t}$, the two models produce next-token distributions:
\begin{equation}
\begin{cases}
p_S^t
=
p_{\theta_S}
\left(
\cdot \mid x,\hat{y}_{<t}
\right), \\[4pt]
p_T^t
=
p_{\theta_T}
\left(
\cdot \mid x,\hat{y}_{<t}
\right),
\end{cases}
\end{equation}

where $\theta_T$ denotes the parameters of the teacher model, and $p_S^t$ and $p_T^t$ represent the student and teacher next-token probability distributions at decoding step $t$, respectively. The student is optimized by minimizing the token-level discrepancy between these two distributions:
\begin{equation}
\mathcal{L}_{\mathrm{OPD}}
=
\frac{1}{|\hat{y}|}
\sum_{t=1}^{|\hat{y}|}
D\left(p_S^t,p_T^t\right),
\end{equation}
where $D(\cdot,\cdot)$ denotes a distribution-level distillation objective. By evaluating the teacher on student-generated prefixes, OPD provides dense supervision directly on the states visited by the current student policy during autoregressive generation.

\begin{figure}
    \centering
    \includegraphics[width=0.98\linewidth]{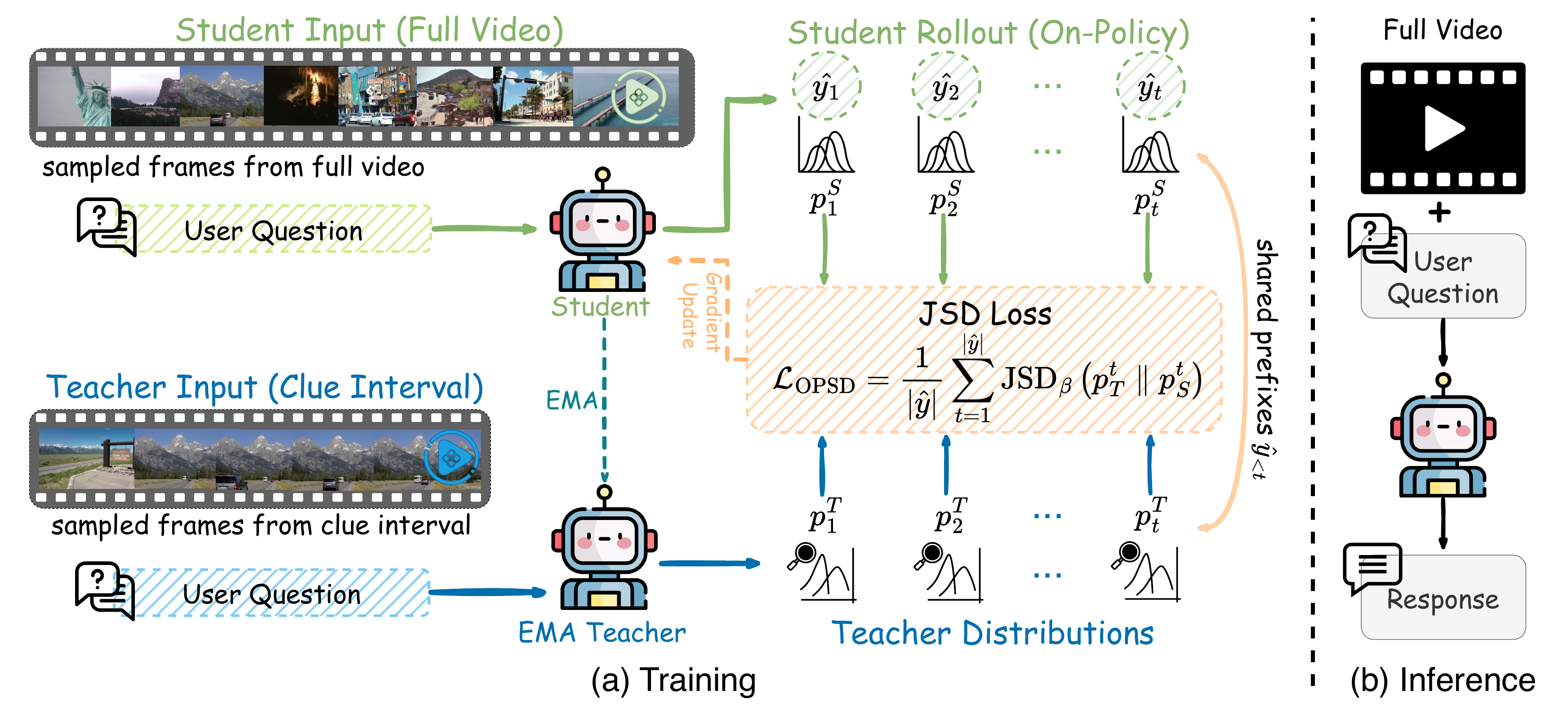}
    \caption{
    Overview of \ours. 
    (a) During training, the student processes frames sampled from the full video and generates an on-policy trajectory, while the EMA self-teacher is conditioned on frames sampled from the corresponding clue interval. 
    Both branches are evaluated along the same student-generated prefixes, and their next-token distributions are aligned using the JSD objective. 
    The teacher parameters are updated through an exponential moving average of the student. 
    (b) At inference time, the clue interval and teacher are removed, and only the student is used with the standard full-video VLM pipeline.
    }
    \label{fig:overall_arch}
\end{figure}

\subsection{Empirical Observation: Temporal Clues Improve Prediction}

\begin{wraptable}{l}{0.52\linewidth}
\vspace{-3mm}
\centering
\small
\setlength{\tabcolsep}{3pt}
\begin{tabular}{lccc}
\hline

\hline
Model & Full Video & Clue Interval & $\Delta$ \\
\hline
Qwen3.5-2B & 45.33 & 58.63 & \textcolor{green!50!black}{+13.30} \\
Qwen3.5-4B & 50.50 & 64.10 & \textcolor{green!50!black}{+13.60} \\
Qwen3.5-9B & 53.47 & 68.43 & \textcolor{green!50!black}{+14.96} \\
\hline
Mean Frames & 1024.00 & 68.15 & \textcolor{red!70!black}{-955.85} \\
\hline

\hline
\end{tabular}
\caption{Accuracy (\%) on 1,000 sampled CG-Bench questions using the full video or the annotated clue interval. $\Delta$ denotes the change from full-video input to clue-interval input. The last row reports the mean number of input frames under the same sampling configuration, with a maximum of 1,024 frames at 2 FPS.}
\label{tab:clue_observation}
\end{wraptable}

Although current VLMs typically process the entire video for long-video question answering, whether the full temporal context is always beneficial for answering a given question remains unclear. To investigate this, we conduct a preliminary comparison on 1,000 randomly sampled multiple-choice questions from CG-Bench~\citep{chen2025cg}. In addition to question-answer annotations, CG-Bench provides clue intervals that contain the visual evidence required to answer each question. This enables us to directly compare model predictions under two visual conditions: the full video and the corresponding annotated clue interval. For each example, we keep the question, answer options, and decoding configuration identical across the two settings.

As shown in Table~\ref{tab:clue_observation}, using the annotated clue interval consistently yields higher accuracy than using the full video across different model scales. Both settings use the same frames-per-second (FPS) and maximum number of frames, while the clue interval is typically much shorter and therefore contains fewer input frames than the full video. The improvement thus does not result from observing more visual information. Instead, it shows that focusing the visual input on the question-relevant temporal interval leads to more effective prediction than sampling frames from the entire video.

This observation naturally motivates our approach: instead of requiring clue annotations at inference time, we use them only during training to construct a privileged teacher and distill its predictions into a student that operates on the full video.

\subsection{\ours: Clue-Privileged On-Policy Self-Distillation}\label{sec:main-method}

Building on the above observation, we introduce \ours, an on-policy self-distillation framework for long-video understanding, which transfers the predictive behavior induced by clue-focused visual inputs to a student that operates on the full video. The detailed architecture is shown in Figure~\ref{fig:overall_arch}. Specifically, the student receives the full-video input, while the teacher is conditioned on the annotated clue interval under the same question and answer options. Rather than relying on a separately pretrained or larger teacher, we maintain the teacher as an exponential moving average (EMA) of the student parameters. As a result, the supervision gap is introduced by the visual context rather than model capacity: the teacher predicts from question-relevant temporal evidence, while the student learns to reproduce such predictions from the full video.

\paragraph{Asymmetric Visual Contexts.}
For each training instance, let $V=\{v_1,\ldots,v_N\}$ denote the sequence of $N$ frames sampled from the full video, and let $V^{\star}=\{v^{\star}_1,\ldots,v^{\star}_M\}$ denote the sequence of $M$ frames sampled from the corresponding annotated clue interval. Both inputs follow the same FPS and maximum-frame constraint. Therefore, we have $M \leq N$, with $M=N$ when both the clue interval and the full video reach the maximum number of sampled frames. The student is conditioned on $V$, while the teacher is conditioned on $V^{\star}$. Both branches receive exactly the same textual input, including the question $q$ and its candidate answer options, such that the only difference between them lies in the visual context.

The student, parameterized by $\theta_S$, first generates an on-policy response $\hat{y}=(\hat{y}_1,\ldots,\hat{y}_T)$ from the full-video input:
\begin{equation}
\hat{y}
\sim
p_{\theta_S}(\cdot \mid V,q),
\end{equation}
where $T$ denotes the number of generated tokens. The generated response is then fixed and used as the shared autoregressive trajectory for both branches. At decoding step $t$, let $\hat{y}_{<t}$ denote the prefix consisting of all tokens generated before step $t$. The student and teacher next-token distributions are respectively defined as:
\begin{equation}
\begin{cases}
p_S^t
=
p_{\theta_S}
\left(
\cdot \mid V,q,\hat{y}_{<t}
\right), \\[4pt]
p_T^t
=
p_{\theta_T}
\left(
\cdot \mid V^{\star},q,\hat{y}_{<t}
\right),
\end{cases}
\end{equation}
where $\theta_T$ denotes the teacher parameters, and $p_S^t$ and $p_T^t$ denote the student and teacher next-token probability distributions at step $t$, respectively.

\paragraph{On-Policy Distribution Alignment.}
Given the student-generated trajectory $\hat{y}$, we align the student and teacher next-token distributions at each decoding step using generalized Jensen--Shannon divergence (JSD). Specifically, the distillation objective is:
\begin{equation}
\mathcal{L}_{\mathrm{OPSD}}=
\frac{1}{|\hat{y}|}
\sum_{t=1}^{|\hat{y}|}
\mathrm{JSD}_{\beta}
\left(
p_T^t
\parallel
p_S^t
\right),
\end{equation}
where $|\hat{y}|$ denotes the length of the student-generated response. For each decoding step, the generalized JSD is defined as:
\begin{equation}
\mathrm{JSD}_{\beta}
\left(
p_T^t
\parallel
p_S^t
\right)
=
\beta
\mathrm{KL}
\left(
p_T^t
\parallel
m^t
\right)
+
(1-\beta)
\mathrm{KL}
\left(
p_S^t
\parallel
m^t
\right),
\end{equation}
where
\begin{equation}
m^t
=
\beta p_T^t
+
(1-\beta)p_S^t
\end{equation}
is the mixture distribution at step $t$, and $\beta$ controls the relative contribution of the two KL terms. During optimization, the teacher distribution is treated as a fixed target and gradients are propagated only through the student branch.

Following Vision-OPD~\citep{yuan2026vision}, we adopt top-$K$ logit distillation to reduce the memory and computation required for full-vocabulary distribution matching. Specifically, we retain the top-$K$ tokens selected from the student distribution, together with the corresponding teacher logits and the remaining tail probability, and compute the distillation objective on this compressed distribution.

By minimizing $\mathcal{L}_{\mathrm{OPSD}}$, the student is encouraged to reproduce the clue-conditioned teacher distribution while operating on the full-video input. Since the shared trajectory is generated by the current student policy, the resulting token-level supervision remains on-policy throughout training.

\paragraph{EMA Self-Teacher.}
We construct the teacher as an EMA of the student rather than introducing a separately pretrained model, following~\citep{zhao2026self}. After each optimization step, the teacher parameters are updated as
\begin{equation}
\theta_T
\leftarrow
(1-\alpha) \cdot \theta_T
+
\alpha \cdot \theta_S,
\end{equation}
where $\alpha$ denotes the EMA update coefficient. The teacher is not optimized by back-propagation; instead, it evolves only through the EMA update and serves as a slowly changing reference for distillation. This design preserves the self-distillation setting while providing a more stable teacher distribution than directly reusing the current student parameters. Combined with the clue-privileged visual input $V^{\star}$, the EMA teacher provides the student with a stable prediction target conditioned on question-relevant temporal evidence.

\paragraph{Training and Inference.}
During training, the updated student parameters are periodically synchronized with the rollout model so that subsequent responses are generated by the current student policy. The clue interval annotations and EMA teacher are used only for training. At inference time, both are discarded, and the student directly processes the full video following the standard VLM inference pipeline, requiring no additional clue information at inference time.

%% file: 4exp.tex
\section{Experiments}

\subsection{Experimental Setting}

\paragraph{Data Construction.}
We construct our training data from CG-Bench~\citep{chen2025cg}, a long-video understanding benchmark that provides question-answer pairs together with annotated temporal clue intervals indicating the evidence required to answer each question. We focus on the multiple-choice questions (MCQs) and retain only samples with valid temporal clue annotations. From the resulting set, we randomly sample 5,000 question-answer instances for training, covering 1,206 distinct long videos. To avoid data leakage, we further verify that the videos used for training do not overlap with those in any of our evaluation benchmarks. This ensures that the reported results reflect generalization to unseen videos rather than memorization of training content.

\paragraph{Models and Baselines.}
We conduct experiments on the Qwen3.5 model family~\citep{qwen35blog}, including Qwen3.5-2B, Qwen3.5-4B, and Qwen3.5-9B, to evaluate the effectiveness of our proposed \ours across different model scales. For each model size, we compare \ours with the corresponding vanilla Qwen3.5 model, as well as models trained on the same data using SFT, GRPO, and standard OPSD. For SFT, the ground-truth answer is directly used as the target response for teacher-forced likelihood optimization. GRPO uses the ground-truth answer to determine the reward for sampled responses, while standard OPSD conditions the teacher on the ground-truth answer as privileged information and keeps the student on the original full-video input. In contrast, \ours uses only temporal clue annotations as privileged supervision and does not use ground-truth answers during training.

\paragraph{Training Details.}
We initialize both the student and EMA teacher from the same pretrained Qwen3.5 checkpoint and perform full-parameter fine-tuning. All models are trained on 8 NVIDIA B200 GPUs with a learning rate of $2\times10^{-6}$, an effective batch size of 32, and up to 300 optimization steps. We adopt a cosine learning-rate schedule with a warmup ratio of $0.03$. We use generalized JSD with $\beta=0.5$, top-$K$ logit distillation with $K=100$, and a distillation temperature of $1.0$. The EMA update coefficient is set to $\alpha=0.05$. The student weights are synchronized with the vLLM rollout engine after every optimization step to ensure on-policy sampling. During rollout, we use a sampling temperature of $1.0$, top-$p=0.95$, and top-$k=20$, with a maximum generation length of 2,048 tokens. Both the student and teacher videos are sampled at 2 FPS with at most 256 frames during training.

\paragraph{Benchmarks.}
We evaluate our method on five video understanding benchmarks: Video-MME~\citep{fu2025video}, LVBench~\citep{wang2025lvbench}, LongVideoBench~\citep{wu2024longvideobench}, MLVU~\citep{zhou2025mlvu}, and MMVU~\citep{zhao2025mmvu}. LVBench, LongVideoBench, and MLVU primarily evaluate long-video understanding over extended temporal contexts, while Video-MME provides a broader evaluation across different video durations. For Video-MME, we report performance on the short, medium, and long subsets under both with- and without-subtitle settings. For MLVU, we report the multiple-choice macro-average (M-Avg) following the standard evaluation protocol. For MMVU, we report the overall benchmark accuracy. Together, these benchmarks cover a diverse range of video durations, temporal reasoning requirements, and question-answering settings.

\paragraph{Inference Details.}
At inference time, the EMA teacher and clue annotations are removed, and the student directly processes the full video. We sample videos at 2 FPS and use benchmark-specific maximum frame budgets: 768, 1,024, and 2,048 frames for the short, medium, and long subsets of Video-MME, respectively; 2,048 frames for LVBench, LongVideoBench, and MLVU; and 1,024 frames for MMVU. Following the official Qwen3.5 recommended configuration for non-thinking inference, we use a sampling temperature of $0.7$, top-$p=0.8$, top-$k=20$, a presence penalty of $1.5$, and a repetition penalty of $1.0$, with a maximum generation length of 2,048 tokens. The same decoding configuration is used across all evaluated models and benchmarks. We repeat each evaluation three times and report the mean performance over the three runs.

\begin{table}[t]
\centering
\small
\setlength{\tabcolsep}{5.5pt}
\renewcommand{\arraystretch}{1.08}

\begin{tabular}{llccccccc}
\hline

\hline
\multirow{2}{*}{Model}
& \multirow{2}{*}{Setting}
& \multicolumn{3}{c}{Video-MME (w/o Subtitles)}
& \multicolumn{3}{c}{Video-MME (w/ Subtitles)}
& \multirow{2}{*}{Average} \\
\cline{3-8}
& & Short & Medium & Long & Short & Medium & Long & \\
\hline

\multirow{5}{*}{Qwen3.5-2B}
& Vanilla
& 72.40 & 57.93 & 46.73 & 75.30 & 65.22 & 57.24 & 62.47 \\
& SFT & 66.15 & 57.00 & 47.33 & 70.48 & 62.56 & 53.92 & 59.57 \\
& GRPO & 71.33 & 61.22 & 50.77  &  75.11 &  66.11 & 58.00 & 63.76  \\
& OPSD & 69.70 & 59.44 & 46.67 & 72.93 & 64.67 & 52.78 & 61.03  \\
& \textbf{\ours}
& \textbf{72.78} & \textbf{62.67} & \textbf{51.33}
& \textbf{75.85} & \textbf{67.78} & \textbf{59.11}
& \textbf{64.92} \\

\hline

\multirow{5}{*}{Qwen3.5-4B}
& Vanilla
& 78.02 & 67.09 & 60.16
& 80.33 & 74.49 & 70.56
& 71.78 \\
& SFT & 76.78 & 68.48 & 58.74 & 79.18 & 75.41 & 67.85 & 71.07 \\
& GRPO & 77.96 & 69.15 & 60.70 & 80.37 & 74.00 & 69.89 &  72.01    \\
& OPSD  & \textbf{80.37} & 69.85 & 61.56 & \textbf{82.45} & 74.63 & 71.19 & 73.34 \\
& \textbf{\ours}
& 80.31 & \textbf{72.09} & \textbf{62.29}
& 82.29 & \textbf{75.65} & \textbf{72.48}
& \textbf{74.19} \\

\hline

\multirow{5}{*}{Qwen3.5-9B}
& Vanilla
& 81.04 & 73.42 & 64.47 & 83.11 & 77.91 & 73.42 & 75.56 \\
& SFT & 78.37 & 72.18 & 61.48 & 81.22 & 77.11 & 70.82 & 73.53 \\
& GRPO & 81.67 & 74.11 & 65.22 & 84.66 & 78.07 & 74.00 & 76.29  \\
& OPSD & 81.82 & 74.52 & 65.41 & 84.59 & 77.92 & 74.65 &  76.49 \\
& \textbf{\ours}
& \textbf{82.30} & \textbf{74.89} & \textbf{66.33}
& \textbf{85.33} & \textbf{78.33} & \textbf{74.67}
& \textbf{76.98} \\

\hline

\hline
\end{tabular}

\caption{Results on Video-MME across different video durations under settings with and without subtitles. Each experiment is repeated three times, and the mean performance is reported.}
\label{tab:videomme-results}
\end{table}

\begin{table}[t]
\centering
\small
\setlength{\tabcolsep}{7pt}
\renewcommand{\arraystretch}{1.08}
\begin{tabular}{llcccc}
\hline

\hline
Model & Setting & MLVU & LVBench & LongVideoBench & MMVU \\
\hline

\multicolumn{6}{c}{\textit{Frontier Proprietary Models}} \\
GPT-5              & Reported & 77.7 & 65.2 & 72.6 & -- \\
Gemini 3 Pro       & Reported & 75.7 & 77.0 & 75.9 & 77.5 \\
Gemini 2.5 Pro     & Reported & --   & 75.7 & 76.8 & 75.8 \\
Gemini 2.5 Flash   & Reported & 75.1 & 64.9 & 73.1 & 70.3 \\
\hline

\addlinespace[2pt]
\multicolumn{6}{c}{\textit{Strong Open-Source VLMs}} \\
GLM-4.1V-9B        & Reported & 56.6 & 44.0 & 65.7 & -- \\
MiniCPM-V-4.5-8B   & Reported & 60.6 & 50.4 & 63.9 & -- \\
Molmo2-8B          & Reported & 60.2 & 52.8 & 67.5 & -- \\
InternVL3.5-8B     & Reported & 53.2 & 43.4 & 62.1 & -- \\
InternVL3.5-38B    & Reported & 77.0 & --   & 65.7 & -- \\
InternVL3-78B      & Reported & 79.5 & --   & 65.7 & -- \\

\midrule
\multicolumn{6}{c}{\textit{Qwen3.5-Based Models}} \\

\multirow{5}{*}{Qwen3.5-2B}
& Vanilla & 59.34 & 48.43 & 51.28 & 55.84 \\
& SFT     &   65.74  &   50.14  &  51.78 &  55.04  \\
& GRPO    &  65.89   &  50.43  &  60.23  &  57.59  \\
& OPSD  & 65.67 & 50.48  & 57.44 & 57.92 \\
& \textbf{\ours} & \textbf{66.28} & \textbf{51.88} & \textbf{61.09} & \textbf{59.25} \\

\hline

\addlinespace[2pt]
\multirow{5}{*}{Qwen3.5-4B}
& Vanilla & 68.45 & 54.45 & 61.58 & 63.84 \\
& SFT     &  71.54  &   55.59  &  58.14  & 63.36  \\
& GRPO    &   75.98 & 59.22  &  64.72  &  65.92   \\
& OPSD  & 75.75 & 59.00 & 64.94  &  65.56\\
& \textbf{\ours} & \textbf{76.68} & \textbf{59.87} & \textbf{66.62} & \textbf{66.99} \\

\hline

\addlinespace[2pt]
\multirow{5}{*}{Qwen3.5-9B}
& Vanilla & 74.91 & 60.31 & 66.72 & 72.96 \\
& SFT     &   73.28  &  57.78   &   59.51  &  69.76   \\
& GRPO    &  80.93   &  60.96  & 68.42  &   73.87  \\
& OPSD & \textbf{81.05} & 61.30 & 67.93 & 73.76 \\
& \textbf{\ours} & 80.27 & \textbf{61.53} & \textbf{68.49} & \textbf{74.88} \\

\hline

\hline
\end{tabular}
\caption{Comparison with frontier proprietary models, strong open-source VLMs, and controlled Qwen3.5-based baselines on long-video understanding benchmarks. External model results are taken from publicly reported evaluations. All Qwen3.5-based results are averaged over three runs.}
\label{tab:main-results-2}
\vspace{-2mm}
\end{table}

\subsection{Experimental Results}
\paragraph{Overall Performance.}
Table~\ref{tab:main-results-2} compares \ours with both publicly reported VLMs and controlled Qwen3.5-based baselines. Across all three model scales, \ours consistently improves over the corresponding vanilla backbones on the four evaluated benchmarks. The gains are particularly pronounced for smaller models. For Qwen3.5-2B, \ours improves MLVU, LVBench, LongVideoBench, and MMVU by 6.94, 3.45, 9.81, and 3.41 points, respectively. Similar improvements are observed for Qwen3.5-4B, with gains of 8.23 points on MLVU and 5.42 points on LVBench over the vanilla model. These results show that clue-privileged self-distillation provides consistent benefits across different long-video reasoning settings and model capacities.

Compared with supervised post-training baselines, \ours also achieves competitive performance despite not using ground-truth answer labels during training. On Qwen3.5-2B and Qwen3.5-4B, \ours outperforms SFT, GRPO, and standard OPSD across all four benchmarks. For Qwen3.5-9B, \ours achieves the best results on LVBench, LongVideoBench, and MMVU, while remaining competitive with OPSD on MLVU. Notably, these improvements are obtained using only temporal clue annotations as privileged supervision, suggesting that question-relevant temporal evidence provides an effective learning signal beyond direct answer supervision. In addition, \ours remains competitive with substantially larger open-source and proprietary VLMs on several benchmarks, while retaining the original Qwen3.5 architecture and standard full-video inference pipeline.

\begin{table}[h]
\centering
\small
\setlength{\tabcolsep}{8pt}
\renewcommand{\arraystretch}{1.08}
\begin{tabular}{lcccc}
\hline

\hline
Divergence Objective & MLVU & LVBench & LongVideoBench & MMVU \\
\hline
Forward KL
($\mathrm{KL}(p_T \| p_S)$)
& 76.21  & 59.14 & 65.84 & 66.41 \\

Reverse KL
($\mathrm{KL}(p_S \| p_T)$)
& 75.86 & 58.93 & 66.08 & 66.62 \\

JSD ($\beta=0.5$)
& \textbf{76.68} & \textbf{59.87} & \textbf{66.62} & \textbf{66.99} \\
\hline

\hline
\end{tabular}

\caption{Effect of different divergence objectives for token-level distribution alignment on Qwen3.5-4B. All other training and inference configurations are kept unchanged.}
\label{tab:divergence-ablation}
\end{table}

\paragraph{Performance across Video Durations.}
Table~\ref{tab:videomme-results} reports detailed results on Video-MME across short, medium, and long videos under both subtitle settings. \ours consistently improves the average performance of Qwen3.5-2B, 4B, and 9B from 62.47, 71.78, and 75.56 to 64.92, 74.19, and 76.98, respectively. The improvements are especially clear on the medium- and long-video subsets, where identifying relevant temporal evidence becomes increasingly important. For example, on Qwen3.5-2B without subtitles, \ours improves the medium- and long-video accuracy from 57.93 and 46.73 to 62.67 and 51.33, respectively. Similar trends hold for larger models and when subtitles are available.

Compared with the supervised baselines, \ours achieves the highest average Video-MME accuracy at all three model scales. In particular, it improves over the strongest competing post-training baseline by 1.16, 0.85, and 0.49 points for Qwen3.5-2B, 4B, and 9B, respectively. The consistent gains across video durations and subtitle settings further indicate that the benefit of clue-privileged training is not tied to a particular evaluation condition.

\subsection{Ablation Studies}

\paragraph{Effect of Divergence Objective.}
We study the effect of different divergence objectives used for token-level distribution matching between the teacher and student. Specifically, we compare forward KL divergence, reverse KL divergence, and Jensen--Shannon divergence (JSD) with $\beta=0.5$. We keep all other training configurations unchanged and conduct this ablation on Qwen3.5-4B only. As shown in Table~\ref{tab:divergence-ablation}, JSD provides the most consistent overall performance across the four benchmarks. Compared with forward and reverse KL, JSD achieves higher accuracy on MLVU, LVBench, and MMVU, while remaining competitive on LongVideoBench. These results suggest that JSD provides a more balanced distribution-matching objective for clue-privileged on-policy self-distillation. We therefore adopt JSD with $\beta=0.5$ as the default divergence objective in all main experiments.

\paragraph{Effect of Privileged Clue Utilization.}
We investigate how temporal clue annotations should be incorporated into the privileged teacher. Besides the default \ours setting, where the teacher directly observes the annotated clue interval, we compare a full-video teacher with ground-truth answers, a full-video teacher with the clue interval provided only as textual side information, and a teacher conditioned on a randomly sampled interval with the same duration. As shown in Table~\ref{tab:clue-utilization}, directly conditioning the teacher on the annotated clue interval achieves the best performance across all four benchmarks. For example, on MLVU, the clue-interval teacher reaches 76.68, compared with 73.66 for the textual-clue variant and 52.32 for the random-interval variant. The large drop with random intervals shows that the gain does not come from using a shorter visual input alone, but from exposing the teacher to question-relevant temporal evidence. Moreover, the clue-interval teacher consistently outperforms the answer-privileged OPSD baseline, including 66.62 vs. 64.94 on LongVideoBench, despite not using ground-truth answer labels. These results support using clue interval as the privileged condition for on-policy self-distillation.

\begin{table}[t]
\centering
\small
\setlength{\tabcolsep}{7pt}
\renewcommand{\arraystretch}{1.08}
\begin{tabular}{lccccc}
\hline

\hline
Teacher Visual Input & Additional Privilege & MLVU & LVBench & LongVideoBench & MMVU \\
\hline

Full Video & GT Answer
& 75.75 & 59.00 & 64.94 & 65.56 \\

Full Video & Textual Clue
& 73.66 & 58.70 & 64.17 & 63.57 \\

Random Interval & -
& 52.32 &  37.85 & 51.78 & 57.44 \\

Clue Interval  & -
& \textbf{76.68} & \textbf{59.87} & \textbf{66.62} & \textbf{66.99} \\

\hline

\hline
\end{tabular}
\caption{Comparison of different privileged teacher constructions for on-policy distillation on Qwen3.5-4B. ``Textual Clue'' provides the annotated temporal interval in the prompt while retaining the full-video visual input. ``Random Interval'' uses a randomly sampled interval with the same duration as the ground-truth clue interval. The clue-interval setting corresponds to the default configuration of \ours.}
\label{tab:clue-utilization}
\end{table}

\paragraph{Sensitivity to Clue Interval Precision.}
We further investigate how sensitive \ours is to the temporal precision of clue annotations.
In practice, precisely identifying the temporal boundaries of question-relevant evidence may require additional annotation effort, while coarser temporal annotations are easier to obtain.
To simulate increasingly imprecise clue annotations, we expand each annotated clue interval by 10, 30, and 60 seconds on both temporal boundaries, respectively, while clipping the expanded interval to the valid video range.
These expanded intervals preserve the original question-relevant evidence while introducing increasing amounts of surrounding temporal context.
We train \ours on Qwen3.5-4B under each setting, with all other training configurations kept unchanged.

As shown in Table~\ref{tab:clue_expansion}, the original clue intervals achieve the best average performance of 67.54, while the expanded settings remain competitive, reaching 66.02, 66.66, and 65.76 under $\pm$10s, $\pm$30s, and $\pm$60s expansion, respectively.
Notably, the degradation is not monotonic with respect to the expansion range: the $\pm$30s setting outperforms $\pm$10s on average and remains close to the original setting on LVBench and MMVU.
Even when the clue interval is expanded by 60 seconds on both sides, the average performance decreases by only 1.78 points.
These results suggest that \ours does not critically depend on highly precise temporal boundaries.
Instead, retaining the question-relevant evidence within the privileged teacher input appears to be more important than tightly localizing its exact temporal extent.

\begin{table}[t]
\centering
\caption{
Sensitivity of \ours to the temporal precision of clue annotations on Qwen3.5-4B.
Each annotated clue interval is expanded by the specified duration on both temporal boundaries, while all other training configurations are kept unchanged.
}

\label{tab:clue_expansion}
\begin{tabular}{lccccc}
\toprule
Teacher Clue Interval & MLVU & LVBench & LongVideoBench & MMVU & Average \\
\midrule
Original          & 76.68 & 59.87 & 66.62 & 66.99 & 67.54 \\
Expanded $\pm$10s & 74.34 & 59.46 & 65.32 & 64.96 & 66.02 \\
Expanded $\pm$30s & 74.09 & 59.78 & 65.92 & 66.83 & 66.66 \\
Expanded $\pm$60s & 73.56 & 59.01 & 64.32 & 66.13 & 65.76 \\
\bottomrule
\end{tabular}
\end{table}


\subsection{Discussion}

The advantage of \ours over SFT, GRPO, and standard OPSD can be attributed to the form of supervision provided during training. SFT directly optimizes the ground-truth answer, but does not explicitly expose which visual content in a long video supports that answer. GRPO performs on-policy optimization, yet its supervision is mainly determined by final-answer correctness and is therefore relatively sparse over the generated trajectory. Standard OPSD provides dense on-policy distribution supervision, but the privileged information is typically centered on the ground-truth answer rather than the visual evidence supporting it. In contrast, \ours conditions the teacher on the question-relevant clue interval, so the supervision is generated from a visual context in which the evidence needed for the current question is directly emphasized.

This distinction also helps explain why answer-privileged OPSD can underperform \ours despite having access to the ground-truth answer. The answer specifies \emph{what} the correct prediction is, but does not indicate \emph{which visual evidence} should support that prediction. By contrast, the clue interval changes the teacher's visual condition itself, allowing its next-token distribution to reflect predictions made from concentrated question-relevant evidence rather than from the full-video context. Our ablations support this interpretation: replacing the clue with a random interval causes a large drop, while providing the clue only as textual side information is also consistently weaker than directly conditioning the teacher on the clue interval. These results suggest that the benefit of \ours comes not from shorter inputs or privileged information alone, but from using question-relevant visual evidence as the privileged condition for on-policy self-distillation.

%% file: 5con.tex
\section{Conclusion}
In this work, we study how temporal clue annotations can be used as privileged supervision for long-video understanding. Our empirical analysis shows that short question-relevant clue intervals can provide more effective visual context than the corresponding full videos, even with fewer input frames. Building on this finding, we introduce \ours, a clue-privileged on-policy self-distillation framework that uses a clue-conditioned self-teacher to supervise a full-video student without ground-truth answer labels. Extensive experiments across multiple benchmarks and Qwen3.5 model scales demonstrate consistent improvements over the original backbones and strong performance against supervised post-training baselines. These results demonstrate the effectiveness of temporal clues as privileged supervision for improving full-video VLMs.

%% file: 6appendix.tex
\appendix

\section{Appendix}

\subsection{Prompts}
We use the official evaluation prompts for most benchmarks whenever available. For completeness and reproducibility, we present the prompt templates used for benchmark evaluation below.

\begin{figure}[H]
    \centering
    \begin{tcolorbox}[
        enhanced,
        colback=green!2!white,
        colframe=green!45!black,
        title={Prompt for Video-MME Evaluation without Subtitles},
        coltitle=white,
        fonttitle=\bfseries,
        arc=3mm,
        boxrule=0.8pt,
        left=10pt,
        right=10pt,
        top=8pt,
        bottom=8pt,
        fontupper=\small\ttfamily
    ]
    Select the best answer to the following multiple-choice question based on the video.
    Respond with only the letter (A, B, C, or D) of the correct option.

    \vspace{2mm}

    Question: \{question\}

    \{choices\}

    The best answer is:
    \end{tcolorbox}

    \caption{Prompt template used for Video-MME evaluation without subtitles.}
    \label{fig:videomme-prompt}
\end{figure}

\begin{figure}[H]
    \centering
    \begin{tcolorbox}[
        enhanced,
        colback=green!2!white,
        colframe=green!45!black,
        title={Prompt for Video-MME Evaluation with Subtitles},
        coltitle=white,
        fonttitle=\bfseries,
        arc=3mm,
        boxrule=0.8pt,
        left=10pt,
        right=10pt,
        top=8pt,
        bottom=8pt,
        fontupper=\small\ttfamily
    ]
    This video's subtitles are listed below:

    \{subtitles\}

    \vspace{2mm}

    Select the best answer to the following multiple-choice question based on the video.
    Respond with only the letter (A, B, C, or D) of the correct option.

    \vspace{2mm}

    Question: \{question\}

    \{choices\}

    \vspace{2mm}

    The best answer is:
    \end{tcolorbox}

    \caption{Prompt template used for Video-MME evaluation with subtitles.}
    \label{fig:videomme-subtitle-prompt}
\end{figure}

\begin{figure}[H]
    \centering
    \begin{tcolorbox}[
        enhanced,
        colback=green!2!white,
        colframe=green!45!black,
        title={Prompt for LVBench Evaluation},
        coltitle=white,
        fonttitle=\bfseries,
        arc=3mm,
        boxrule=0.8pt,
        left=10pt,
        right=10pt,
        top=8pt,
        bottom=8pt,
        fontupper=\small\ttfamily
    ]
    Carefully watch the video and pay attention to the cause and sequence of events,
    the detail and movement of objects and the action and pose of persons.

    \vspace{2mm}

    Based on your observations, select the best option that accurately addresses the question.

    \vspace{2mm}

    Question: \{question\}

    \{choices\}

    \vspace{2mm}

    Answer with the option's letter from the given choices directly and only give the best option.
    \end{tcolorbox}

    \caption{Prompt template used for LVBench evaluation.}
    \label{fig:lvbench-prompt}
\end{figure}

\begin{figure}[H]
    \centering
    \begin{tcolorbox}[
        enhanced,
        colback=green!2!white,
        colframe=green!45!black,
        title={Prompt for LongVideoBench Evaluation},
        coltitle=white,
        fonttitle=\bfseries,
        arc=3mm,
        boxrule=0.8pt,
        left=10pt,
        right=10pt,
        top=8pt,
        bottom=8pt,
        fontupper=\small\ttfamily
    ]
    \{question\}

    \{options\}

    \vspace{2mm}

    Answer with the option's letter from the given choices directly.
    \end{tcolorbox}

    \caption{Prompt template used for LongVideoBench evaluation.}
    \label{fig:longvideobench-prompt}
\end{figure}

\begin{figure}[H]
    \centering
    \begin{tcolorbox}[
        enhanced,
        colback=green!2!white,
        colframe=green!45!black,
        title={Prompt for MLVU Evaluation},
        coltitle=white,
        fonttitle=\bfseries,
        arc=3mm,
        boxrule=0.8pt,
        left=10pt,
        right=10pt,
        top=8pt,
        bottom=8pt,
        fontupper=\small\ttfamily
    ]
    Carefully watch this video and pay attention to every detail.
    Based on your observations, select the best option that accurately addresses the question.

    \vspace{2mm}

    Question: \{question\}
    
    Options: \{options\}

    \vspace{2mm}

    Only give the best option.
    \end{tcolorbox}

    \caption{Prompt template used for MLVU evaluation.}
    \label{fig:mlvu-prompt}
\end{figure}

\begin{figure}[H]
    \centering
    \begin{tcolorbox}[
        enhanced,
        colback=green!2!white,
        colframe=green!45!black,
        title={Prompt for MMVU Evaluation},
        coltitle=white,
        fonttitle=\bfseries,
        arc=3mm,
        boxrule=0.8pt,
        left=10pt,
        right=10pt,
        top=8pt,
        bottom=8pt,
        fontupper=\small\ttfamily
    ]
    \{question\}

    \{options\}

    \vspace{2mm}

    Answer with the option's letter from the given choices directly.
    \end{tcolorbox}

    \caption{Prompt template used for MMVU evaluation.}
    \label{fig:mmvu-prompt}
\end{figure}

\subsection{Algorithm}

Algorithm~\ref{alg:clue-opsd} summarizes the training procedure of \ours. For each training instance, the student first generates an on-policy response from the full-video input. The same student-generated prefixes are then used to evaluate both the full-video student and the clue-conditioned EMA teacher. Their next-token distributions are aligned through the JSD distillation objective, with gradients applied only to the student. After each optimization step, the teacher is updated as an exponential moving average of the latest student parameters, and the rollout model is synchronized accordingly.

\begin{algorithm}[t]
\caption{Training Procedure of \ours}
\label{alg:clue-opsd}
\begin{algorithmic}[1]

\Require Training set
$\mathcal{D}=\{(\mathcal{V},q,I^\star)\}$,
pretrained VLM parameters $\theta$,
EMA coefficient $\alpha$,
JSD coefficient $\beta$,
top-$K$ value $K$

\Ensure Trained student parameters $\theta_S$

\State Initialize student and teacher:
$\theta_S \leftarrow \theta$,
$\theta_T \leftarrow \theta$

\State Initialize rollout model with $\theta_S$

\For{each optimization step}

    \State Sample a mini-batch $\mathcal{B}$ from $\mathcal{D}$
    \State Initialize batch loss $\mathcal{L} \leftarrow 0$

    \For{each training instance $(\mathcal{V},q,I^\star)\in\mathcal{B}$}

        \State Sample full-video frames $V$ from $\mathcal{V}$
        \State Sample clue frames $V^\star$ from the annotated clue interval $I^\star$

        \State Generate an on-policy response from the full-video student:
        \Statex \hspace{\algorithmicindent}
        $\hat{y}\sim p_{\theta_S}(\cdot\mid V,q)$

        \For{$t=1,\ldots,|\hat{y}|$}

            \State Compute the student distribution:
            \Statex \hspace{\algorithmicindent}
            $p_S^t
            =
            p_{\theta_S}
            (\cdot\mid V,q,\hat{y}_{<t})$

            \State Compute the teacher distribution on the same prefix:
            \Statex \hspace{\algorithmicindent}
            $p_T^t
            =
            p_{\theta_T}
            (\cdot\mid V^\star,q,\hat{y}_{<t})$

            \State Construct top-$K$ compressed student and teacher distributions

        \EndFor

        \State Accumulate the instance-level distillation loss:
        \Statex \hspace{\algorithmicindent}
        $\displaystyle
        \mathcal{L}
        \leftarrow
        \mathcal{L}
        +
        \frac{1}{|\hat{y}|}
        \sum_{t=1}^{|\hat{y}|}
        \mathrm{JSD}_{\beta}
        \left(
        p_T^t
        \parallel
        p_S^t
        \right)$

    \EndFor

    \State Average the loss over the mini-batch:
    \Statex \hspace{\algorithmicindent}
    $\mathcal{L}
    \leftarrow
    \mathcal{L}/|\mathcal{B}|$

    \State Update the student parameters $\theta_S$ by back-propagating $\mathcal{L}$

    \State Update the EMA teacher:
    \Statex \hspace{\algorithmicindent}
    $\theta_T
    \leftarrow
    (1-\alpha)\theta_T
    +
    \alpha\theta_S$

    \State Synchronize the rollout model with the updated student $\theta_S$

\EndFor

\State \Return $\theta_S$

\end{algorithmic}
\end{algorithm}